\documentclass[times,twocolumn,final,authoryear]{elsarticle}

\usepackage{prletters}
\usepackage{framed,multirow}

\usepackage{amssymb}
\usepackage{latexsym}

\usepackage{graphicx}
\usepackage{comment}
\usepackage{amsmath} 

\usepackage[table]{xcolor}
\definecolor{s3}{HTML}{57bb8a}
\definecolor{s5}{HTML}{e67c73}
\definecolor{s40}{HTML}{e88b83}
\definecolor{s94}{HTML}{acdec6}
\definecolor{s118}{HTML}{a1d9be}
\definecolor{s63}{HTML}{c8e9d9}
\definecolor{s117}{HTML}{90d2b2}
\definecolor{s32}{HTML}{b5e1cc}
\definecolor{s35}{HTML}{62c092}
\definecolor{s34}{HTML}{5cbd8d}
\definecolor{s155}{HTML}{eca099}
\definecolor{s184}{HTML}{e6f5ee}
\definecolor{s7}{HTML}{daf0e5}
\definecolor{s13}{HTML}{f9e2e0}
\definecolor{s106}{HTML}{e9f6f0}
\definecolor{s192}{HTML}{f6fcf9}
\definecolor{s23}{HTML}{92d3b3}
\definecolor{s9}{HTML}{d8efe4}
\definecolor{s191}{HTML}{f7d6d3}
\definecolor{s152}{HTML}{f2beba}
\definecolor{s56}{HTML}{f7d5d3}
\definecolor{s120}{HTML}{ebf7f1}
\definecolor{s18}{HTML}{f8dedc}
\definecolor{s82}{HTML}{6bc398}
\definecolor{s128}{HTML}{e7f5ee}
\definecolor{s173}{HTML}{c7e8d8}
\definecolor{s39}{HTML}{efaca6}
\definecolor{s15}{HTML}{fdf8f8}
\definecolor{s75}{HTML}{9ad6b9}
\definecolor{s146}{HTML}{f1b6b1}
\definecolor{s193}{HTML}{fdf9f8}
\definecolor{s198}{HTML}{f2bfbb}
\definecolor{s88}{HTML}{fbecea}
\definecolor{s142}{HTML}{88cfac}
\definecolor{s4}{HTML}{ffffff}
\definecolor{s178}{HTML}{d9f0e4}
\definecolor{s59}{HTML}{e7857d}
\definecolor{s182}{HTML}{99d6b8}
\definecolor{s189}{HTML}{eaf7f1}
\definecolor{s51}{HTML}{8bd0ae}
\definecolor{s60}{HTML}{fdf7f6}
\definecolor{s125}{HTML}{fdf6f6}
\definecolor{s38}{HTML}{f4c9c6}
\definecolor{s84}{HTML}{6cc499}
\definecolor{s95}{HTML}{58bc8b}
\definecolor{s83}{HTML}{6ac398}
\definecolor{s35}{HTML}{62c092}
\definecolor{s107}{HTML}{fdf9f9}
\definecolor{s115}{HTML}{f6d1ce}
\definecolor{s44}{HTML}{f4cac6}
\definecolor{s68}{HTML}{e67e75}
\definecolor{s135}{HTML}{eaf7f0}
\definecolor{s108}{HTML}{64c193}
\definecolor{s22}{HTML}{daf1e6}
\definecolor{s52}{HTML}{89cfad}
\definecolor{s139}{HTML}{e8f6ef}
\definecolor{s140}{HTML}{b0dfc8}
\definecolor{s179}{HTML}{f0b5b0}
\definecolor{s147}{HTML}{69c397}
\definecolor{s63}{HTML}{c8e9d9}
\definecolor{s117}{HTML}{90d2b2}
\definecolor{s164}{HTML}{e7847b}
\definecolor{s190}{HTML}{e4f4ed}
\definecolor{s126}{HTML}{f8dcd9}
\definecolor{s111}{HTML}{c6e8d7}
\definecolor{s180}{HTML}{fae6e4}
\definecolor{s32}{HTML}{b5e1cc}
\definecolor{s160}{HTML}{f7d8d6}
\definecolor{s2}{HTML}{f3c1bd}
\definecolor{s3}{HTML}{57bb8a}
\definecolor{s166}{HTML}{e98e87}
\definecolor{s70}{HTML}{c5e8d7}
\definecolor{s113}{HTML}{fbedec}
\definecolor{s177}{HTML}{e5f5ed}
\definecolor{s72}{HTML}{fefdfd}
\definecolor{s49}{HTML}{f8dcda}
\definecolor{s94}{HTML}{acdec6}
\definecolor{s57}{HTML}{efaea9}
\definecolor{s100}{HTML}{b6e2cc}
\definecolor{s122}{HTML}{f7d7d5}
\definecolor{s30}{HTML}{e8877f}
\definecolor{s194}{HTML}{fdfefe}
\definecolor{s21}{HTML}{f5cac7}
\definecolor{s105}{HTML}{f7d5d2}
\definecolor{s141}{HTML}{a6dbc1}
\definecolor{s123}{HTML}{f8dddb}
\definecolor{s11}{HTML}{d0ecdf}
\definecolor{s16}{HTML}{a4dbc0}
\definecolor{s159}{HTML}{eda39d}
\definecolor{s29}{HTML}{f3c2be}
\definecolor{s76}{HTML}{eb9890}
\definecolor{s162}{HTML}{caeada}
\definecolor{s34}{HTML}{5cbd8d}
\definecolor{s81}{HTML}{65c194}
\definecolor{s5}{HTML}{e67c73}
\definecolor{s67}{HTML}{fdf7f7}
\definecolor{s158}{HTML}{fae7e6}
\definecolor{s134}{HTML}{e0f3ea}
\definecolor{s157}{HTML}{efaba5}
\definecolor{s27}{HTML}{fbeceb}
\definecolor{s98}{HTML}{fcf0ef}
\definecolor{s165}{HTML}{eb9a93}
\definecolor{s133}{HTML}{fcf4f3}
\definecolor{s20}{HTML}{fcf3f3}
\definecolor{s92}{HTML}{5ebe8f}
\definecolor{s12}{HTML}{e98c84}
\definecolor{s43}{HTML}{ccebdc}
\definecolor{s102}{HTML}{fefbfb}
\definecolor{s90}{HTML}{6fc59b}
\definecolor{s187}{HTML}{f8dad7}
\definecolor{s110}{HTML}{a5dbc1}
\definecolor{s175}{HTML}{f5cecb}
\definecolor{s62}{HTML}{eef9f3}
\definecolor{s46}{HTML}{fae9e7}
\definecolor{s112}{HTML}{fae8e7}
\definecolor{s25}{HTML}{5dbe8f}
\definecolor{s181}{HTML}{f5cfcc}
\definecolor{s132}{HTML}{f4c5c2}
\definecolor{s37}{HTML}{fcefee}
\definecolor{s171}{HTML}{bfe5d2}
\definecolor{s8}{HTML}{eb968f}
\definecolor{s167}{HTML}{fbfefd}
\definecolor{s64}{HTML}{75c79f}
\definecolor{s163}{HTML}{75c89f}
\definecolor{s19}{HTML}{f3faf7}
\definecolor{s161}{HTML}{e3f4ec}
\definecolor{s153}{HTML}{fefbfa}
\definecolor{s148}{HTML}{f7d9d7}
\definecolor{s17}{HTML}{edf8f3}
\definecolor{s138}{HTML}{eeaaa4}
\definecolor{s74}{HTML}{7ac9a2}
\definecolor{s151}{HTML}{9ed8bb}
\definecolor{s168}{HTML}{f9fdfb}
\definecolor{s119}{HTML}{9fd9bd}
\definecolor{s136}{HTML}{f2faf6}
\definecolor{s69}{HTML}{fae9e8}
\definecolor{s196}{HTML}{95d4b5}
\definecolor{s137}{HTML}{f4fbf7}
\definecolor{s26}{HTML}{d4eee1}
\definecolor{s31}{HTML}{f4c7c3}
\definecolor{s170}{HTML}{f9e1df}
\definecolor{s50}{HTML}{a2dabf}
\definecolor{s116}{HTML}{d6efe2}
\definecolor{s91}{HTML}{5dbe8e}
\definecolor{s195}{HTML}{fcfefd}
\definecolor{s96}{HTML}{addec6}
\definecolor{s154}{HTML}{c3e7d5}
\definecolor{s114}{HTML}{def2e8}
\definecolor{s121}{HTML}{f8dbd9}
\definecolor{s24}{HTML}{7fcba6}
\definecolor{s85}{HTML}{68c296}
\definecolor{s176}{HTML}{fae5e4}
\definecolor{s155}{HTML}{eca099}
\definecolor{s55}{HTML}{dff2e9}
\definecolor{s183}{HTML}{c2e7d5}
\definecolor{s93}{HTML}{6dc49a}
\definecolor{s89}{HTML}{fefefe}
\definecolor{s87}{HTML}{f7d9d6}
\definecolor{s42}{HTML}{a7dcc2}
\definecolor{s65}{HTML}{fdf5f5}
\definecolor{s145}{HTML}{fbeeed}
\definecolor{s48}{HTML}{e98f87}
\definecolor{s109}{HTML}{69c296}
\definecolor{s186}{HTML}{f6d4d1}
\definecolor{s127}{HTML}{f5fbf8}
\definecolor{s129}{HTML}{ecf8f2}
\definecolor{s131}{HTML}{f2bcb8}
\definecolor{s40}{HTML}{e88b83}
\definecolor{s41}{HTML}{93d4b4}
\definecolor{s118}{HTML}{a1d9be}
\definecolor{s66}{HTML}{dbf1e6}
\definecolor{s199}{HTML}{f9e4e2}
\definecolor{s61}{HTML}{f3c4c0}
\definecolor{s172}{HTML}{7ecba5}
\definecolor{s104}{HTML}{f1b8b3}
\definecolor{s79}{HTML}{fcf2f1}
\definecolor{s6}{HTML}{8fd2b1}
\definecolor{s197}{HTML}{eda59f}
\definecolor{s144}{HTML}{efafa9}
\definecolor{s71}{HTML}{a0d9bd}
\definecolor{s14}{HTML}{a4dac0}
\definecolor{s174}{HTML}{78c8a1}
\definecolor{s97}{HTML}{83cda9}
\definecolor{s185}{HTML}{ccebdb}
\definecolor{s130}{HTML}{f5cbc7}
\definecolor{s80}{HTML}{f2bbb6}
\definecolor{s47}{HTML}{f8dbd8}
\definecolor{s150}{HTML}{feffff}
\definecolor{s86}{HTML}{eea6a0}
\definecolor{s10}{HTML}{f9e3e1}
\definecolor{s149}{HTML}{cfecde}
\definecolor{s36}{HTML}{e2f4eb}
\definecolor{s58}{HTML}{e99088}
\definecolor{s169}{HTML}{ddf1e7}
\definecolor{s54}{HTML}{fdf4f4}
\definecolor{s101}{HTML}{fefaf9}
\definecolor{s28}{HTML}{f6d3d0}
\definecolor{s45}{HTML}{b8e2ce}
\definecolor{s73}{HTML}{b6e2cd}
\definecolor{s99}{HTML}{f8fcfa}
\definecolor{s53}{HTML}{a9dcc3}
\definecolor{s188}{HTML}{94d4b5}
\definecolor{s124}{HTML}{fdf8f7}
\definecolor{s143}{HTML}{e7847c}
\definecolor{s77}{HTML}{fae8e6}
\definecolor{s156}{HTML}{e88a82}
\definecolor{s33}{HTML}{70c59c}
\definecolor{s78}{HTML}{fef9f9}

\usepackage{booktabs}
\usepackage{multirow}
\usepackage{subfiles}
\usepackage{url}

\newcommand{\norm}[1]{\left\lVert#1\right\rVert}

\usepackage{fancyhdr}
\journal{Pattern Recognition Letters}

\begin{document}

\ifpreprint
  \setcounter{page}{1}
\else
  \setcounter{page}{1}
\fi

\begin{frontmatter}

\title{Disentangling Image Distortions in Deep Feature Space}

\author[1]{Simone \snm{Bianco}} 
\author[1]{Luigi \snm{Celona}\corref{cor1}}
\cortext[cor1]{Corresponding author:
  Tel.: +39-02-6448-7871;}
\ead{luigi.celona@unimib.it}
\author[1]{Paolo \snm{Napoletano}}
\address[1]{University of Milano - Bicocca -- DISCo, viale Sarca, 336, 20126, Milano, Italy
}

\received{1 May 2013}
\finalform{10 May 2013}
\accepted{13 May 2013}
\availableonline{15 May 2013}
\communicated{S. Sarkar}

\begin{abstract}
Previous literature suggests that perceptual similarity is an emergent property shared across deep visual representations. Experiments conducted on a dataset of human-judged image distortions have proven that deep features outperform classic perceptual metrics. In this work we take a further step in the direction of a broader understanding of such property by analyzing the capability of deep visual representations to intrinsically characterize different types of image distortions. To this end, we firstly generate a number of synthetically distorted images and then we analyze the features extracted by different layers of different Deep Neural Networks. We observe that a dimension-reduced representation of the features extracted from a given layer permits to efficiently separate types of distortions in the feature space. Moreover, each network layer exhibits a different ability to separate between different types of distortions, and this ability varies according to the network architecture. Finally, we evaluate the exploitation of features taken from the layer that better separates image distortions for: i) reduced-reference image quality assessment, and ii) distortion types and severity levels characterization on both single and multiple distortion databases. Results achieved on both tasks suggest that deep visual representations can be unsupervisedly employed to efficiently characterize various image distortions.
\end{abstract}


\end{frontmatter}



\begin{figure*}
    \centering
    \includegraphics[width=.75\textwidth]{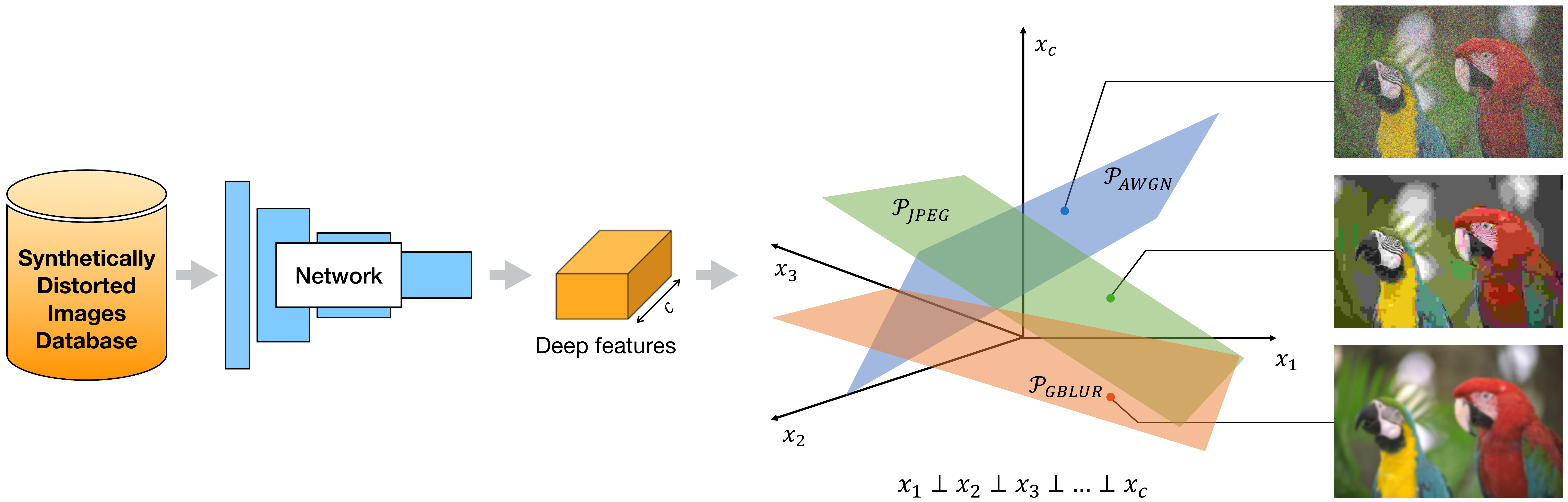}
    \caption{Are deep features intrinsically able to characterize image distortions? The deep features extracted from a specific layer of a network trained for a recognition task, intrinsically divide the deep feature space in such a way that visual representations corresponding to different types of distortions lay on different planes.}
    \label{fig:teaser}
\end{figure*}

%
\section{Motivation}
The networks trained to discriminate a wide variety of categories contained within the images allow to obtain rich representations that can be reused for other tasks in which it is necessary to carry out a semantic analysis of the image. Deep features extracted from common networks trained for recognition tasks, such as ImageNet challenge \cite{russakovsky2015imagenet}, have been demonstrated to be very effective for transfer learning in many tasks \cite{sharif2014cnn}.
Moreover, what makes pretrained networks good for transfer learning is not directly connected to the amount of data but probably to the network architecture itself \cite{huh2016makes}. Literature has also highlighted another surprising property of pretrained networks: visual deep representation can be adopted as metric for perceptual similarity. This property was firstly experienced for feature inversion \cite{mahendran2015understanding}, feature visualization \cite{simonyan2013deep,yosinski2015understanding}, texture synthesis and style transfer \cite{gatys2015texture,gatys2015neural}. The most remarkable paper in this direction demonstrates that high-level features from a pretrained network on ImageNet can be efficiently employed as a perceptual loss to train feed-forward transformation networks for image transformation tasks \cite{johnson2016perceptual}. A recent paper \cite{zhang2018unreasonable} tried to answer to the following questions: i) ``how perceptual are these so-called perceptual losses?''; ii) ``what elements are critical for their success?''. The findings of this paper suggest that perceptual similarity is an emergent property shared across deep visual representations that outperform, by a large margin, classic perceptual metrics such as SSIM~\cite{wang2004image} for image distortion characterization.  This property has been assessed on a large-scale highly diverse dataset of perceptual judgments specially collected for this scope. The dataset included images with different kinds of image distortions: traditional, obtained by performing basic low-level image editing operations; CNN-based distortions, obtained by randomly varying parameters, etc. Although the paper provides a comprehensive demonstration that deep features catch perceptual properties of the images better than traditional metrics, further understanding should be provided in the direction of explaining how deep features characterize image distortions.

We want here to go one step further. What we would really like to understand is: \textbf{are deep networks able to effectively characterize image distortions? If so, is this capability different on the basis of network architectures? Which layer provides the representation better disentangling distortions? Can dimensional reduction techniques help to strengthen this property?} In this paper, we conduct a comprehensive study of deep features extracted across different networks and compare them in terms of distortions separability indices. This is motivated by the fact that we assume that deep features representing images affected by the same type of distortions should be similar and so they might form clusters. We find that deep features are able to intrinsically disentangle image distortions and that this property is not equally revealed across networks and layers.

The main contributions of this paper are as follows:
\begin{itemize}
    \item We conduct a comprehensive analysis of deep features for several networks in order to understand which one is the most effective for image distortion characterization.
    \item We show that deep features obtaining a high distortion separability index can obtain comparable results with respect to reduced-reference image quality assessment methods on four widely used benchmark databases for the distortion characterization task.
    \item We demonstrate that the evaluated property might be exploited for the recognition of image distortion types and severity levels on single and multiple distortion databases.
\end{itemize}
\section{Data}
To evaluate how visual features deal with different types and severity of image distortions, we create our own database, that differently from available ones, has a larger and controlled number of severity levels. In particular, we generate 783 synthetically distorted images by applying three mainstream distortion types to the 29 reference images of the LIVE database \cite{sheikh2005live}: additive white Gaussian noise (AWGN), Gaussian blur (GBlur), and JPEG compression (JPEG). More in detail, each reference image is corrupted by applying each one of the following operations:
\begin{itemize}
    \item AWGN: The noise is generated from a standard normal distribution of zero mean and standard deviation $\sigma_N$ and then added to each color channel;
    \item GBlur: A Gaussian kernel with standard deviation $\sigma_G$ is applied for blurring with a square kernel window each of the three planes using the function \texttt{gaussian\_filter} of the Scipy.ndimage \cite{2019arXiv190710121V} package;
    \item JPEG: JPEG compressed images are obtained by varying the quality parameter ($Q$) of \texttt{save} function of the Pillow library\footnote{Pillow is a fork of the Python Imaging Library (PIL): \url{https://pillow.readthedocs.io/en/stable/index.html}.}(whose range is from 1 to 100, with 100 representing the best quality), which indicates the degree of the JPEG compression algorithm.
\end{itemize}
For each type of image distortion we consider the following values which cause several levels of severity: 

\noindent $\sigma_N =(0.03, 0.06, 0.09, 0.13, 0.18, 0.24, 0.31, 0.50, 1.89)$, \\
\noindent $\sigma_G=(0.62, 0.82, 0.95, 1.13, 1.42, 1.65, 2.17, 3.54, 13.00)$, \\
\noindent $Q=(80, 60, 45, 30, 20, 15, 10, 5, 2)$.

\begin{figure}
    \centering
    \includegraphics[width=\columnwidth]{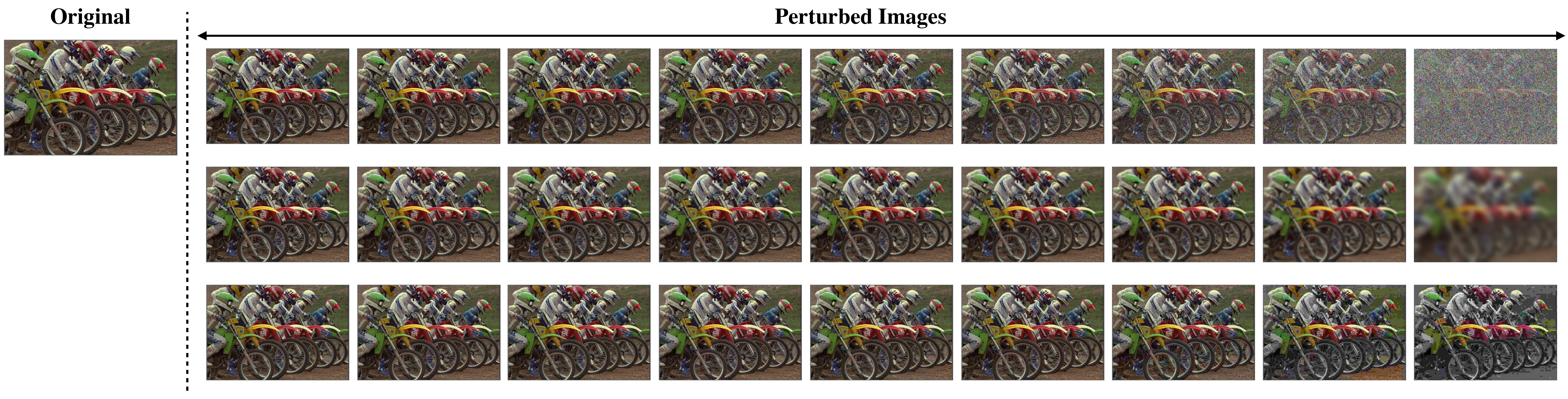}
    \caption{Samples from the generated database. The original image and corresponding perturbed versions with increasing distortion severity from the left to the right. First row shows images affected by additive white Gaussian noise (AWGN), the second row contains images corrupted by Gaussian blur (GBlur), finally the last row reports images with different JPEG compression levels (JPEG).}
    \label{fig:samples}
\end{figure}

Values are chosen in such a way that distorted images are perceptually separable from each other.  
Each type and severity level of image distortions are applied to each reference image. Figure \ref{fig:samples} shows for a given reference image the corresponding synthetically generated distorted images.

\section{Disentangling Image Distortions}
\label{sec:deep_features2cluster_metrics}
\begin{figure}
    \centering
    \includegraphics[width=\columnwidth]{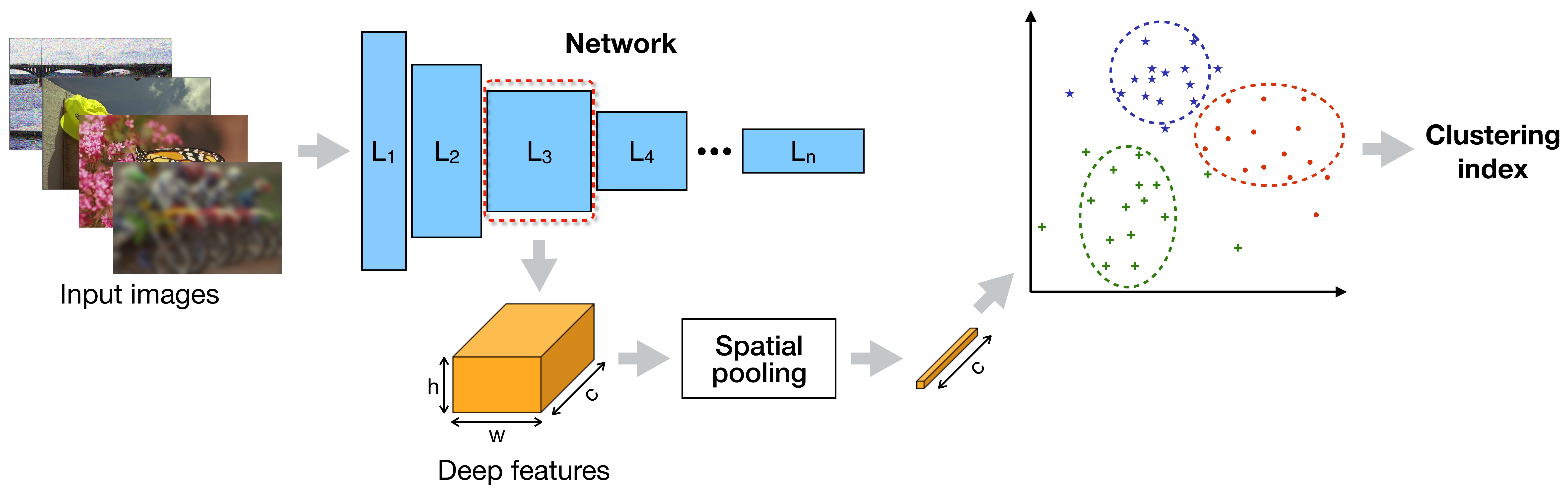}
    \caption{Clusters evaluation. Each full-size image of the generated database is fed into a network, the deep features are obtained by using the activations of a specific layer, so the average spatial pooling is applied and finally, we measure the quality of the clusters that group the images affected by the same type of distortion.}
    \label{fig:cluster-measure}
\end{figure}

The intuition here is that a visual deep feature space, suitably defined, permits to efficiently disentangle image distortion types. To pursue this idea, we firstly build a deep feature space by aggregating features extracted from a single layer of a network. Secondly, we evaluate how much such a space effectively permits to separate among the three types of distortions by computing separability indices of the feature space. Figure \ref{fig:cluster-measure} illustrates the entire pipeline for disentangling image distortions.
To evaluate how much this property of the feature space depends on the architecture of the network, we evaluate several architectures.
\paragraph{Network architectures} We consider the AlexNet \cite{krizhevsky2012imagenet}, Inception-v3 \cite{szegedy2016rethinking}, ResNet-50 \cite{he2016deep}, SqueezeNet-v1.1 \cite{iandola2016squeezenet}, and VGG-16 \cite{simonyan2014very} architectures. We use the \texttt{conv1-conv5} layers from AlexNet, which is a popular deep CNN model widely applied in computer vision and includes computations that are loosely matched to the human visual cortex, such as pooling and local response normalization. We extract features from the three convolution layers, namely \texttt{2a3x3}, \texttt{3b1x1}, \texttt{4a3x3}, and all the inception layers (called \texttt{mixed}) of the Inception-v3 network, which introduced the use of factorized convolutions by using asymmetric convolutions. For ResNet-50, we extract features for \texttt{conv1} and subsequent macro-residual blocks (named \texttt{layer}). We evaluate the SqueezeNet-v1.1 architecture by using the first \texttt{conv1} layer and all the \texttt{fire} layers activations. Finally, we take the \texttt{conv} layers of VGG-16, which is usually used as a perceptual metric in image generation.
\paragraph{Deep Feature Space} Given a network level $l$, producing a deep feature block of size $h_l\times w_l\times c_l$, and an image taken from the synthetically distorted dataset, we calculate a deep visual representation by averaging features across spatial dimensions, thus obtaining a vector of size $1 \times c_l$ (see Fig. \ref{fig:cluster-measure}). The images are used at their original sizes so as not to mask artifacts.
\paragraph{Data separation} All the $1\times c_l$ vectors extracted from all the images of the synthetically distorted dataset are grouped by relying on the three types of distortions: AWGN, GBlur and JPEG. Goodness of the resulting clusters is then evaluated using several separability metrics.
\section{Cluster separability indices}
\label{sec:metrics}
To evaluate the quality of the clusters that are formed using the deep features computed for images affected by different types and levels of distortions, we consider a combination of three internal cluster validity indices \cite{arbelaitz2013extensive,Craenendonck2015UsingIV}, which rely only on properties intrinsic of the structure of clusters and their relations to each other, namely: the Calinski-Harabasz index \cite{calinski1974dendrite}, the Davies-Bouldin index~\cite{davies1979cluster}, and the Silhouette index \cite{rousseeuw1987silhouettes}.
%
\paragraph{Calinski-Harabasz index} The Calinski-Harabasz (CH) index \cite{calinski1974dendrite}, also known as Variance Ratio Criterion, estimates the clusters goodness in terms of ratio between the \textit{between-clusters} variance ($SS_B$) and the \textit{within-clusters} variance ($SS_W$). It is defined as follows:
\begin{align}
    CH = \frac{SS_B}{SS_W}\times \frac{N-K}{K-1},
\end{align}
where $N$ is the number of samples, and $K =3$ represents the number of clusters (which are AWGN, GBlur and JPEG). $SS_W$ is specified as $SS_W = \sum_{k=1}^{K}{\sum_{x\in C_k}{\norm{x-G_k}^2}}$, given $C_k$ the set of samples in the cluster $k$, and $G_k$ the center of the cluster $k$. Finally, $SS_B = \sum_{k=1}^{K}{n_k\norm{G_k-G}^2}$, where $n_k$ is the number of points in the cluster $k$, and $G$ is the center of the whole dataset. The score is higher when elements from the same clusters are quite close and clusters itself are well separated.
\paragraph{Davies-Bouldin index} The Davies-Bouldin (DB) index \cite{davies1979cluster} measures the average similarity between clusters and is based on the dispersion measure of a cluster ($\delta_k$) and the cluster dissimilarity measure ($\Delta_{kk'}$). The dispersion measure is the mean distance of the points belonging to cluster $C_k$ to their center $G_k$, $\delta_k = \frac{1}{n_k} \sum_{x \in C_k}{\norm{x-G_k}}$, while $\Delta_{kk'}=\norm{G_{k'}-G_{k}}$ is the distance between the centroids $G_{k'}$ and $G_{k}$ of clusters $C_{k'}$ and $C_k$. The DB is then defined as:
\begin{align}
    \qquad DB=\frac{1}{K}\sum^{K}_{k=1}{\max_{k'\neq k}\left(\frac{\delta_k+\delta_{k'}}{\Delta_{kk'}}\right)}.
\end{align}
Zero is the lowest possible score, values closer to zero indicate a better separation between data.

%
\paragraph{Silhouette index} The Silhouette index (S) \cite{rousseeuw1987silhouettes} is defined as the mean of the silhouette widths for each sample and is formalized as follows:
\begin{align}
    S=\frac{1}{K}\sum_{k=1}^{K}{\frac{1}{n_k}\sum_{x \in C_k}{s(x)}},
\end{align}
where the silhouette width $s(x)$ for each sample is computed as $s(x) = \frac{b(x)-a(x)}{\max(a(x),b(x))}$. $a(x)$ is the mean distance between the sample $x$, such that $x \in C_k$, and the set of samples in the cluster it belongs to, $C_k$. $b(x)$ represents the mean dissimilarity of a sample $x$ with respect to the nearest cluster, namely $b(x)=\min_{k'\neq k}\frac{1}{n_{k'}}\sum_{y\in C_{k'}}{d(x,y)}$. Higher values of the silhouette index denotes a better quality of clusters.
\paragraph{Combination of separability indices}
To understand which layer of a network provides a better separation of clusters we combine the previous indices into a single one as follows: first, we individually perform a min-max normalization in the range [0,1] of the scores obtained for each index by taking the maximum and minimum values for that index over all the layers $l \in L$ of all the network architectures $n \in N$ considered:
\begin{align}
x^\prime(l)=\frac{x(l)-\min\limits_{l \in L, n \in N} x(l_n)}{\max\limits_{l \in L, n \in N} x(l_n) - \min\limits_{l \in L, n \in N}x(l_n)}
\end{align}
with $x=\left\{CH, DB, S\right\}$. Then we estimate the overall distortions separability index $DSI(l)$ for each network layer $l$ in this way:
\begin{align}
    DSI(l) = \frac{1}{3}\left[ CH^\prime(l)+\left(1-DB^\prime(l)\right)+S^\prime(l) \right]
\end{align}
The resulting $DSI(l)$ score ranges between 0 and 1 and the higher is the value the better is the separability of the clusters.
\begin{figure}
    \centering
    \includegraphics[width=.75\columnwidth]{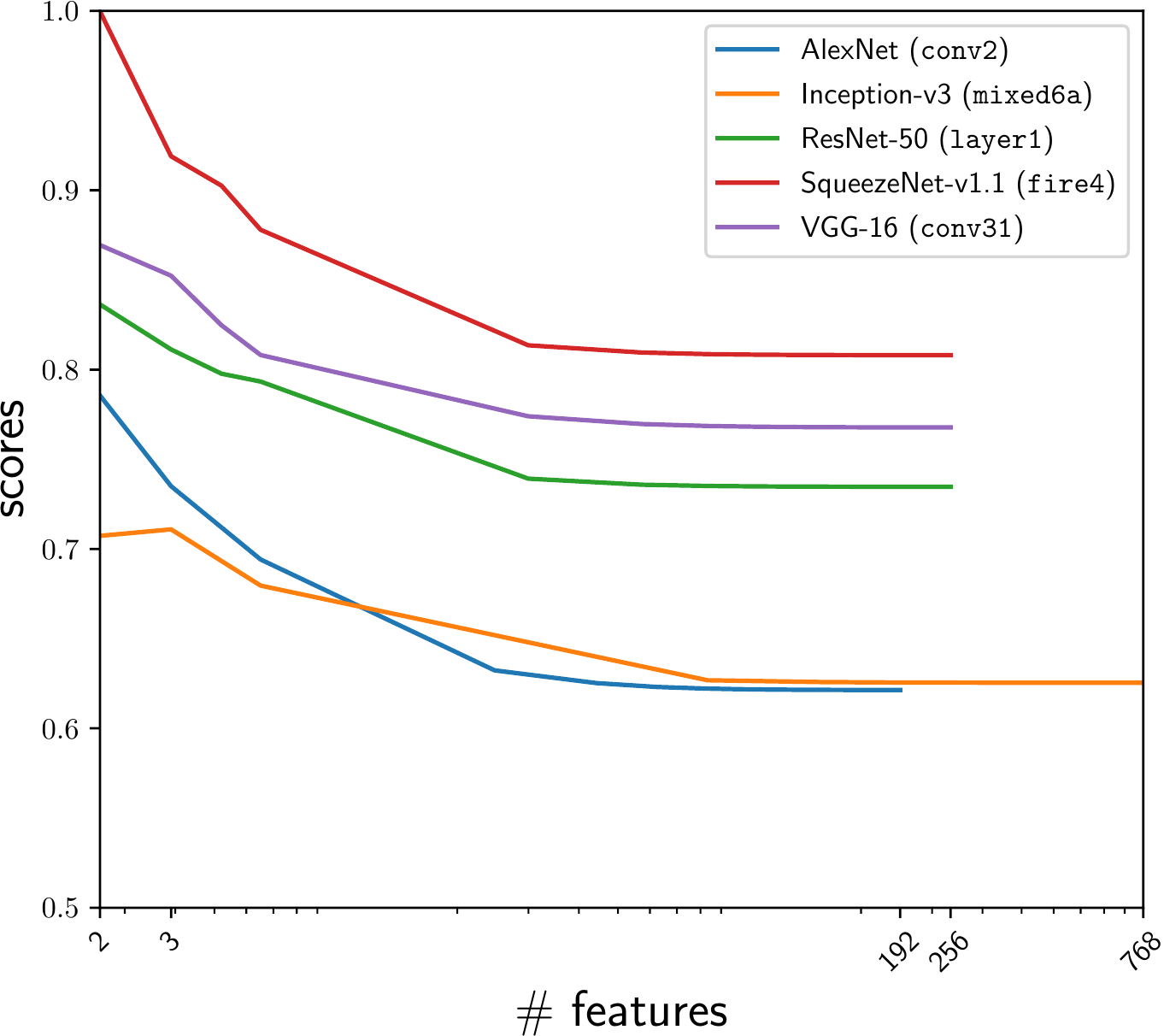}
    \caption{Distortions separability indices obtained by varying the number of features in deep feature vectors using principal component analysis (PCA) for the layer achieving the best distortions separability index of each network considered.}
    \label{fig:cluster_metrics_pca}
\end{figure}
\section{Analysis}
In this section we describe the experiments we conducted to verify the effectiveness of deep features for image distortion characterization. We implement our experiments using the PyTorch framework \cite{paszke2017automatic} and by exploiting the pretrained networks on ImageNet contained in the Torchvision package.
\paragraph{Are the deep features extracted from different networks equally capable of characterizing the various types of distortions? Are all the layers of the network equally effective to characterize distortions?} Table \ref{tab:cluster-results-pretrained} reports the distortions separability index $DSI(l)$ achieved on each layer $l$ of the different networks, pretrained on Imagenet, we considered. Low scores (in red) denote that clusters can not be well separated, while high scores (in green) denote the opposite. The resulting rank of layers per network indicates that the deep features extracted from the first layer get almost always the worst position. We obtain that the layer for each network achieving the best clustering index is the following: \texttt{conv2} for AlexNet, \texttt{mixed6a} of the Inception-v3 network, \texttt{layer1} in ResNet-50, the layer \texttt{fire4} of SqueezeNet-v1.1, and \texttt{conv31} for the VGG-16 network. Among all the best layers, the \texttt{fire4} layer of SqueezeNet-v1.1 is the best. Previous results confirm our intuition that some visual representations corresponding to different types of distortions distribute in well-separated regions of the deep feature space.
\paragraph{Does dimensionality reduction of deep features improve image distortion characterization?} Figure \ref{fig:cluster_metrics_pca} shows the distortions separability index obtained by varying the number of features thanks to the use of principal component analysis (PCA) \cite{wold1987principal}. For each network, we consider the index for the layer that attained the best score in the previous analysis. We can see that by reducing the dimensionality of the features, the distortions separability index increases for all the networks and more in detail it reaches the peak value for 2-dimensional feature vectors. The only network showing a different trend, i.e. lower distortions separability index for bidimensional vectors, is Inception-v3.
\paragraph{Does the capability of discriminating distortions belong to the network architecture or it depends on the network weights?} In this experiment images are encoded by using deep features extracted from randomly initialized networks. For statistical significance we compute the final distortions separability index for each layer by averaging the scores obtained for 100 network re-initializations. Resulting scores for each layer are definitely lower than the ones obtained in the previous experiment indicating that the degree of separability of images affected by diverse distortions is limited. The ranks of layers per network are completely different from the ones obtained in the previous experiment, but given that the standard deviation of distortions separability indices among layers is small, we can not consider this aspect significant.
\paragraph{Can the removal of semantics from the visual representation of the image emphasize the distortion encoding?} In this experiment we want to understand if the separability property of clusters in the deep feature space is influenced by the content of the images. For doing that, we subtract in channel dimension the vector of features, of size $1 \times c_l$, of the reference image to the feature vector of the distorted image. 
The results of this experiment indicate that the semantics does not affect the rank of the layers for AlexNet, VGG-16, as well as for ResNet-50 which, however, presents an increase in the distortions separability index for the best layer. This may be due to the presence of skip connections that carry low-level information from previous layers. Both Inception-v3 and SqueezeNet-v1.1 obtain a different rank of layers with also lower scores for the distortions separability index.
\begin{table*}
    \centering
    \caption{Distortions separability indices for each layer of the considered networks pretrained on ImageNet. The indices are calculated by combining three internal cluster validity indices and are included in the interval [0,1], in which the highest scores (in green) indicate a high separability of the types of distortions, while the low scores (in red) represent levels not able to separate the clusters that are formed to discriminate the different distortions.}
    \label{tab:cluster-results-pretrained}
    \resizebox{.9\textwidth}{!}{
    \begin{tabular}{l|ccccccccccccccc}
        \toprule
        Network & \multicolumn{14}{c}{Layer} \\ \midrule
        \multirow{2}{*}{AlexNet} & \texttt{conv1} & \texttt{conv2} & \texttt{conv3} & \texttt{conv4} & \texttt{conv5} & \multicolumn{9}{c}{} \\ \vspace{.5em}
         & {\cellcolor[HTML]{daf0e5}0.5748} & \cellcolor{s3}0.6214 & 0.5614 & {\cellcolor[HTML]{eb968f}0.5108} & \cellcolor{s5}0.4978 &  &  &  &  &  &  &  &  &  & \\
        \multirow{2}{*}{Inception-v3} & \texttt{2a3x3} & \texttt{3b1x1} & \texttt{4a3x3} & \texttt{mixed5b} & \texttt{mixed5c} & \texttt{mixed5d} & \texttt{mixed6a} & \texttt{mixed6b} & \texttt{mixed6c} & \texttt{mixed6d} & \texttt{mixed6e} & \texttt{mixed7a} & \texttt{mixed7b} & \texttt{mixed7c} \\ \vspace{.5em}
         & \cellcolor{s5}0.4387 & {\cellcolor[HTML]{f6d1ce}0.5284} & {\cellcolor[HTML]{fdf4f4}0.5652} &  {\cellcolor[HTML]{d6efe2}0.5979} & \cellcolor{s63}0.6052 & \cellcolor{s117}0.6347 & \cellcolor{s3}0.6641 & \cellcolor{s118}0.6254 & {\cellcolor[HTML]{9fd9bd}0.6264} & {\cellcolor[HTML]{ebf7f1}0.5867} & {\cellcolor[HTML]{f8dbd9}0.5388} & {\cellcolor[HTML]{f7d7d5}0.5348} & {\cellcolor[HTML]{f8dddb}0.5412} & {\cellcolor[HTML]{fbeceb}0.5564} & \\
        \multirow{2}{*}{ResNet-50} & \texttt{conv1} & \texttt{layer1} & \texttt{layer2} & \texttt{layer3} & \texttt{layer4} &  &  &  &  &  &  &  &  & \\ \vspace{.5em}
         & \cellcolor{s5}0.4551 & \cellcolor{s3}0.7268 & \cellcolor{s94}0.5930 & 0.4615 & 0.4615 &  &  &  &  &  &  &  &  &  & \\
        \multirow{2}{*}{SqueezeNet-v1.1} & \texttt{conv1} & \texttt{fire1} & \texttt{fire2} & \texttt{fire3} & \texttt{fire4} & \texttt{fire5} & \texttt{fire6} & \texttt{fire7} & \texttt{fire8} &  &  &  &  & \\ \vspace{.5em}
        & \cellcolor{s5}0.5186 & {\cellcolor[HTML]{fae5e4}0.7002} & {\cellcolor[HTML]{e5f5ed}0.7535} & {\cellcolor[HTML]{ddf1e7}0.7567} & \cellcolor{s3}0.8081 & {\cellcolor[HTML]{d9f0e4}0.7583} & {\cellcolor[HTML]{f8dddb}0.6863} & {\cellcolor[HTML]{f0b5b0}0.6174} & 0.7433 &  &  &  &  &  & \\
        \multirow{2}{*}{VGG-16} & \texttt{conv11} & \texttt{conv12} & \texttt{conv21} & \texttt{conv22} & \texttt{conv31} & \texttt{conv32} & \texttt{conv33} & \texttt{conv41} & \texttt{conv42} & \texttt{conv43} & \texttt{conv51} & \texttt{conv52} & \texttt{conv53} & \\
         & {\cellcolor[HTML]{f4c7c3}0.5190} & \cellcolor{s32}0.6650 & {\cellcolor[HTML]{70c59c}0.7408} & \cellcolor{s34}0.7634 & \cellcolor{s3}0.7678 & \cellcolor{s35}0.7559 & {\cellcolor[HTML]{e2f4eb}0.6153} & 0.5830 & {\cellcolor[HTML]{fcefee}0.5651} & {\cellcolor[HTML]{f4c9c6}0.5218} & {\cellcolor[HTML]{efaca6}0.4879} & \cellcolor{s40}0.4506 & \cellcolor{s5}0.4327 &  & \\ \bottomrule
    \end{tabular}}
\end{table*}
\section{Applications}
We demonstrate usefulness of the investigated property by experimenting on two quality assessment tasks, namely: reduced-reference image quality assessment (RR-IQA) on four public databases and distortion recognition on single and multiple distortion databases. 
\subsection{Reduced-Reference Image Quality Assessment}
We conduct experiments on the reduced-reference image quality assessment (RR-IQA) task \cite{wang2006modern,wang2011reduced} in which we do not try to estimate the exact quality score but we verify that the rank between the images with different distortions is respected. First we encode both the distorted image and its reference one by using the activations of the layers we demonstrated achieve the best distortions separability index, i.e. layers of pretrained networks on ImageNet. Then we measure the pairwise Euclidean distance between the feature vectors of the distorted image and its reference image, and finally we estimate the correlation between this distance and the database ground-truth. We evaluate the method on four common image databases, which are LIVE \cite{sheikh2005live}, CSIQ \cite{larson2010most}, TID2008 \cite{ponomarenko2009tid2008}, TID2013 \cite{ponomarenko2013color}. Information about these four databases is summarized in Table \ref{tab:iqa-dbs}.
\begin{table}
    \centering
    \caption{Description of the four image quality databases used for the reduced-reference image quality assessment.}
    \label{tab:iqa-dbs}
    \resizebox{\columnwidth}{!}{
    \begin{tabular}{l|cccc}
    \toprule
        Databases & \multicolumn{1}{c}{\begin{tabular}[c]{@{}c@{}}Reference\\ images\end{tabular}} & \multicolumn{1}{c}{\begin{tabular}[c]{@{}c@{}}Distorted\\ images\end{tabular}} & \multicolumn{1}{c}{\begin{tabular}[c]{@{}c@{}}Distortion\\ types\end{tabular}} & \multicolumn{1}{c}{\begin{tabular}[c]{@{}c@{}}Distortion\\ levels\end{tabular}} \\ \midrule
        CSIQ \cite{larson2010most} & 30 & 886 & 6 & 4-5 \\
        LIVE \cite{sheikh2005live} & 29 & 779 & 5 & 7-8 \\
        TID2008 \cite{ponomarenko2009tid2008} & 25 & 1,700 & 17 & 4 \\
        TID2013 \cite{ponomarenko2013color} & 25 & 3,000 & 24 & 5 \\ \bottomrule
    \end{tabular}}
\end{table}

Table \ref{tab:rr-iqa-results-all-distortions} reports the median and the mean SROCC and PLCC for each database and network considered. The \texttt{fire4} layer of the SqueezeNet-v1.1 network provides deep visual representations correlating well with human observers for CSIQ and LIVE databases. Instead, for both the TID2008 and TID2013 databases  the \texttt{conv2} layer of AlexNet is the one achieving the highest correlation values.

Table \ref{tab:rr-iqa-results-comparison-sota} compares state of the art methods with our best solutions on each one of the three types of distortions we considered. First of all, we want to highlight that even without any training, the performance of our solutions attains comparable results with respect to methods explicitly developed to face the image quality assessment task. 
In particular, the average SROCC across all distortions and datasets ranks S4RR \cite{zhang2017reduced} as the best method with an average correlation of 0.9549, followed by our proposed method using AlexNet with an average correlation of 0.9238.

Moreover, considering each distortion type separately, on AWGN the proposed method with AlexNet obtains the highest average SROCC equal to 0.9550 followed by S4RR with an average SROCC of 0.9522; on JPEG S4RR obtains the highest average SROCC of 0.9563, while the proposed method with AlexNet obtains the second highest average SROCC of 0.9435; on GBlur instead S4RR obtains the highest average correlation of 0.9563, with SSIM obtaining the second highest correlation of 0.9238.
\begin{table}
\caption{Median and mean SROCC and PLCC values across 100 train-test random splits on CSIQ, LIVE, TID2008, and TID2013 databases for the considered networks.}
\label{tab:rr-iqa-results-all-distortions}
\centering
\resizebox{.8\columnwidth}{!}{
\begin{tabular}{ll|cccc}
\toprule
\multirow{2}{*}{Database} & \multirow{2}{*}{Network} & \multicolumn{2}{c}{Median} & \multicolumn{2}{c}{Mean}\\
 & & SROCC & PLCC & SROCC & PLCC \\ \midrule
\multirow{5}{*}{CSIQ} & AlexNet & 0.9300 & 0.9050 & 0.9284 & 0.9003 \\
 & Inception-v3 & 0.9120 & 0.9106 & 0.9105 & 0.9081 \\
 & ResNet-50 & 0.9228 & \textbf{0.9146} & 0.9225 & \textbf{0.9111} \\
 & SqueezeNet-v1.1 & \textbf{0.9301} & 0.9072 & \textbf{0.9300} & 0.9055 \\
 & VGG-16 & 0.8898 & 0.8547 & 0.8793 & 0.8394 \\  \midrule
\multirow{5}{*}{LIVE} & AlexNet & 0.8926 & 0.7840 & 0.8892 & 0.7832 \\
 & Inception-v3 & 0.8938 & 0.8884 & 0.8914 & 0.8865 \\
 & ResNet-50 & 0.9240 & 0.9047 & 0.9231 & 0.9047 \\
 & SqueezeNet-v1.1 & \textbf{0.9337} & \textbf{0.9149} & \textbf{0.9314} & \textbf{0.9050} \\
 & VGG-16 & 0.8919 & 0.7791 & 0.8891 & 0.7784 \\  \midrule
\multirow{5}{*}{TID2008} & AlexNet & \textbf{0.9261} & \textbf{0.8671} & \textbf{0.9125} & 0.8359 \\
 & Inception-v3 & 0.7441 & 0.7507 & 0.7422 & 0.7481 \\
 & ResNet-50 & 0.8457 & 0.8572 & 0.8463 & \textbf{0.8589} \\
 & SqueezeNet-v1.1 & 0.8271 & 0.8162 & 0.8068 & 0.7924 \\
 & VGG-16 & 0.8268 & 0.8304 & 0.8150 & 0.8012 \\ \midrule
\multirow{5}{*}{TID2013} & AlexNet & \textbf{0.9490} & \textbf{0.9184} & \textbf{0.9365} & \textbf{0.8885} \\
 & Inception-v3 & 0.7865 & 0.7995 & 0.7881 & 0.7975 \\
 & ResNet-50 & 0.8688 & 0.8785 & 0.8684 & 0.8767 \\
 & SqueezeNet-v1.1 & 0.8558 & 0.8367 & 0.8490 & 0.8202 \\ 
 & VGG-16 & 0.8629 & 0.8707 & 0.8584 & 0.8382 \\ \bottomrule
\end{tabular}}
\end{table}
\begin{table}
    \centering
    \caption{Comparison with state-of-the-art image quality assessment (IQA) methods in terms of mean SROCC values across 100 train-test random splits on CSIQ, LIVE, TID2008, and TID2013 databases. Full-reference IQA (FF-IQA) metrics are in \textit{italic}. The best result for each line is reported in bold, while the best result using the proposed method on deep features is underlined.}
    \label{tab:rr-iqa-results-comparison-sota}
    \resizebox{\columnwidth}{!}{\begin{tabular}{lc|cc|cc|cc}
    \toprule
        \multirow{3}{*}{Database} & \multirow{3}{*}{Dist. type} & \multicolumn{2}{c}{FR-IQA} & \multicolumn{2}{|c}{RR-IQA} & \multicolumn{2}{|c}{Deep features} \\
         & & \textit{PSNR} & \textit{SSIM} & REDLOG & S4RR  & \textbf{AlexNet} & \textbf{Sqz.Net-v1.1} \\
        \midrule
        \multirow{3}{*}{CSIQ} & AWGN & 0.933 & \textbf{0.953} & 0.865 & 0.941 & \underline{\textbf{0.953}} & 0.898 \\
            & GBlur & 0.942 & 0.930 & 0.948 & \textbf{0.958} & 0.897 & \underline{0.935} \\
            & JPEG & 0.896 & 0.921 & {0.936} & \textbf{0.959} & \underline{0.923} & 0.912 \\ \midrule
        \multirow{3}{*}{LIVE} & AWGN & \textbf{0.985} & 0.979 & 0.939 & 0.965 & \underline{0.984} & 0.943 \\
            & GBlur & 0.793 & 0.874 & 0.906 & \textbf{0.944} & 0.832 & \underline{0.912} \\
            & JPEG & 0.901 & 0.934 & {0.950} & \textbf{0.975} & \underline{0.945} & 0.921 \\ \midrule
        \multirow{3}{*}{TID2008} & AWGN & 0.903 & 0.865 & 0.829 & \textbf{0.939} & \underline{0.936} & 0.774 \\
            & GBlur & 0.875 & {0.931} & 0.861 & \textbf{0.952} & 0.846 & \underline{0.864} \\
            & JPEG & 0.903 & 0.916 & 0.905 & \textbf{0.949} & \underline{0.945} & 0.835 \\ \midrule
        \multirow{3}{*}{TID2013} & AWGN & 0.921 & 0.894 & 0.875 & \textbf{0.964} & \underline{0.947} & 0.827 \\
            & GBlur & 0.920 & {0.960} & 0.895 & \textbf{0.971} & \underline{0.916} & 0.897 \\
            & JPEG & 0.930 & 0.927 & 0.897 & {0.942} & \underline{\textbf{0.961}} & 0.867 \\
        \bottomrule
    \end{tabular}}
\end{table}
\subsection{Distortion Recognition}
We carry out experiments designed to evaluate the performance of the investigated property for distortion type recognition and severity level estimation on both single- and multiple- distortion databases. Distorted images are represented as previously explained by extracting deep features from the layer achieving the best distortions separability index and subsequently a \textit{k}-NN classifier is used by considering $k=3$ and $k=9$. We consider accuracy as the metric for quantifying the effectiveness of the method. We run 100 iterations of train-val split and calculate the average performance for the final results.
\paragraph{Single distortion databases}
We conduct experiments on four image quality assessment databases, namely CSIQ, LIVE, TID2008 and TID2013, for each of the networks considered in the previous analysis. Table \ref{tab:rr-iqa-dist-reco} provides the results for two different tasks: 
in the first half of the table the task considered is the distortion recognition; in the second half the task is the combination of distortion recognition and severity level estimation tasks. For the second task only the CSIQ, TID2008 and TID2013 databases are considered since for the LIVE database the information about the distortion level is not available.
For both the two tasks, in accordance with the results obtained in Figure \ref{fig:cluster_metrics_pca}, the best results are obtained by SqueezeNet-v1.1 and VGG-16. 
In particular, for the first task the top performance of 92.2\% accuracy is achieved by SqueezeNet-v1.1 on TID2008 with the 3-NN classifier. On the second task the top performance of 76.5\% accuracy is achieved by VGG-16 on TID2008.
\paragraph{Multiple distortion databases} Experiments have been performed on LIVE multiple distortions (LIVEMD), which is a multiply-distorted image databases. The LIVEMD \cite{jayaraman2012objective} database consists of two subsets attained by synthetically distorting 15 reference images. The first subset includes images obtained by applying the combination of GBlur and JPEG at different severity levels, while the second subset includes images distorted by GBlur and AWGN. There are 450 distorted images in total. Table \ref{tab:rr-iqa-multidist-reco} reports results for the two tasks faced: distortion type recognition and combination of distortion severity and type recognition.
On both tasks the best results are obtained by ResNet-50 with an accuracy of 90.9\% and 48.5\% respectively. Since the difference in performance between the two tasks is very large, we further analyze the results of both tasks. 

\begin{figure}
    \centering
    \includegraphics[width=.6\columnwidth]{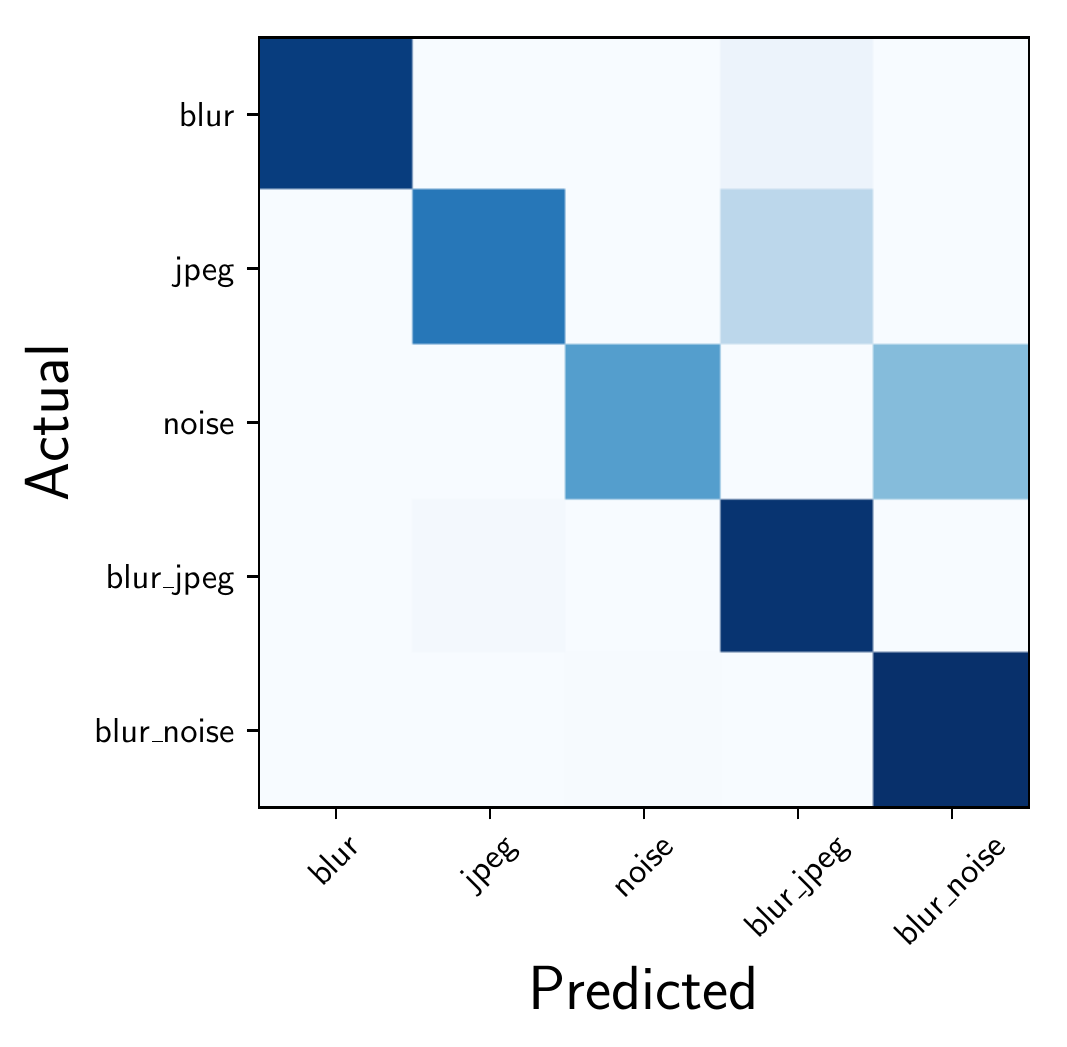}
    \caption{Mean confusion matrix across 100 train-test splits using ResNet-50 deep features and 9-NN classifier on the LIVEMD database only for distortion type recognition.}
    \label{fig:cm_resnet50_multidisttypereco}
\end{figure}

Figure \ref{fig:cm_resnet50_multidisttypereco} reports the average confusion matrix across 100 train-test splits using  ResNet-50 deep features on the LIVEMD database for multiple distortion type recognition. From the confusion matrix it is possible to see how GBlur is perfectly recognized when alone and when in combination with both JPEG and AWGN. JPEG and AWGN alone are sometimes confused with their respective combination with Gblur.

\begin{figure}
    \centering
    \includegraphics[width=.75\columnwidth]{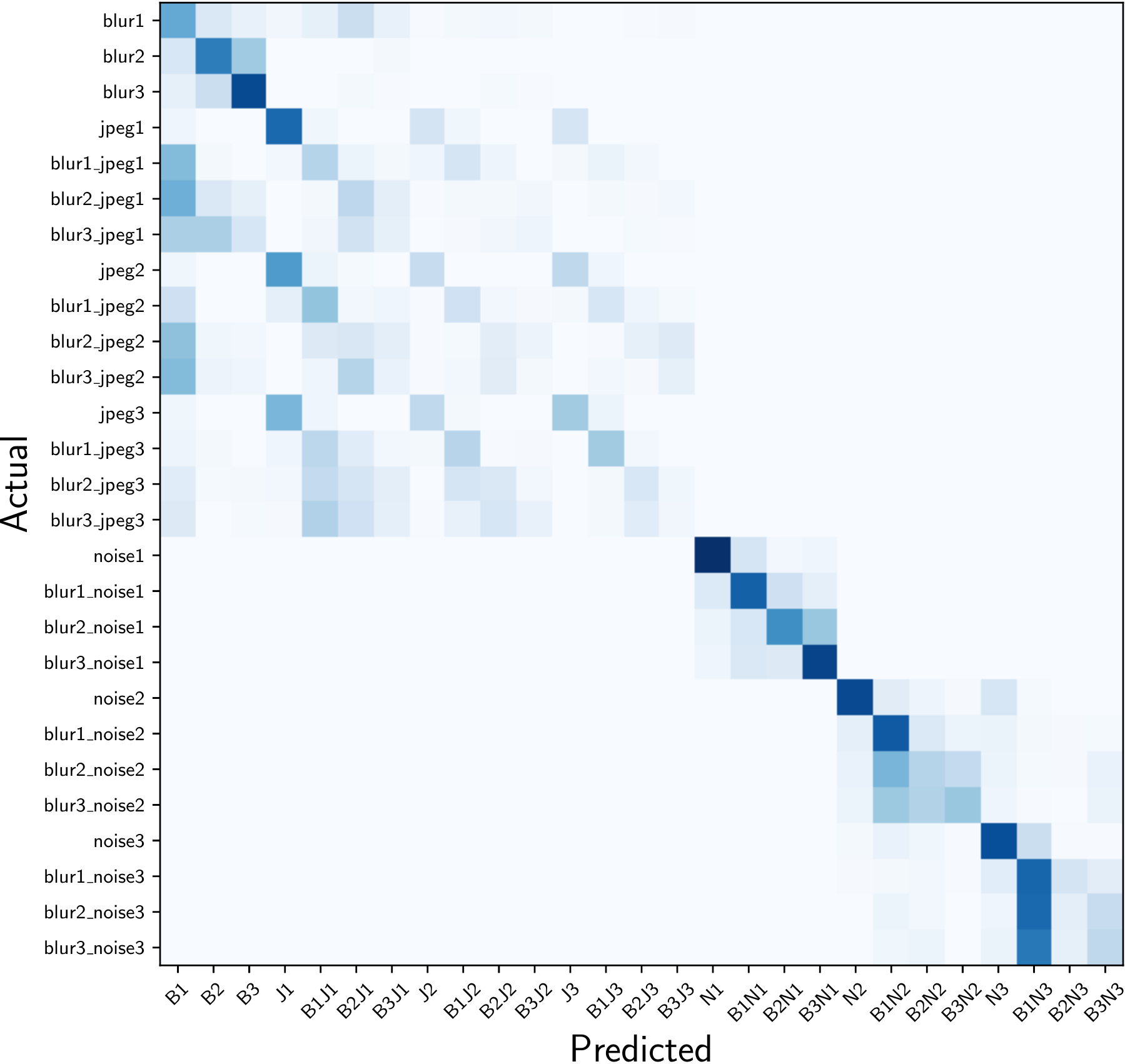}
    \caption{Mean confusion matrix across 100 train-test splits using ResNet-50 deep features and 9-NN classifier on the LIVEMD database for multiple distortion types and severity level recognition.}
    \label{fig:cm_resnet50_multidist}
\end{figure}

Figure \ref{fig:cm_resnet50_multidist} reports the average confusion matrix across 100 train-test splits using  ResNet-50 deep features on the LIVEMD database for multiple distortion types and severity level recognition. The two subsets are reported in the same confusion matrix and are displayed in the top-left and bottom-right quarters respectively. In the first subset (i.e. combination of GBlur and JPEG) it is possible to see that the network is able to discriminate very well among blur levels and level-1 of JPEG, while show some larger confusion in discriminating among higher levels of JPEG. 
In the second subset (i.e. combination of GBlur and AWGN) it is possible to see how the confusion matrix is much more concentrated along the diagonal. In particular we can observe how the network is more able to discriminate among blur levels than noise levels. 
Moreover GBlur has a form of masking on both JPEG and AWGN, in accordance to the results of the distortion type recognition experiment of Figure \ref{fig:cm_resnet50_multidisttypereco}.
\begin{table}
    \caption{Mean classification accuracy across 100 train-test random splits on CSIQ, LIVE, TID2008 and TID2013 databases for two tasks: only for distortion recognition (table top); distortion recognition and severity level estimation (table bottom).}
    \label{tab:rr-iqa-dist-reco}
    \centering
    \resizebox{\columnwidth}{!}{
    \begin{tabular}{l|ccccccccccccc}
    \toprule
        \multirow{2}{*}{Database} & \multicolumn{10}{c}{Classification accuracy (\%)} \\
         & \multicolumn{2}{c}{AlexNet} & \multicolumn{2}{c}{Inception-v3} & \multicolumn{2}{c}{ResNet-50} & \multicolumn{2}{c}{SqueezeNet-v1.1} & \multicolumn{2}{c}{VGG-16} \\
         & 3-NN & 9-NN & 3-NN & 9-NN & 3-NN & 9-NN & 3-NN & 9-NN & 3-NN & 9-NN \\ \midrule
        CSIQ & 70.9 & 74.8 & 59.6 & 59.7 & 74.5 & 76.8 & 78.3 & \textbf{79.6} & 77.1 & 77.9 \\
        LIVE & 84.6 & 86.4 & 69.6 & 71.2 & 84.6 & 86.0 & 83.6 & 83.8 & 85.7 & \textbf{86.5} \\
        TID2008 & 89.0 & 90.2 & 73.0 & 75.0 & 85.6 & 86.9 & \textbf{92.2} & 91.7 & 89.5 & 90.3 \\
        TID2013 & 78.0 & 80.4 & 65.4 & 68.8 & 79.3 & 82.0 & \textbf{85.4} & 85.1 & 82.4 & 85.1 \\
        \midrule
        CSIQ & 23.4 & 28.8 & 19.7 & 21.6 & 40.4 & 45.1 & 41.5 & \textbf{45.7} & 42.5 & 43.6 \\
        TID2008 & 43.8 & 46.9 & 36.5 & 38.5 & 58.2 & 57.7 & 58.0 & 65.0 & \textbf{76.5} & 74.9 \\
        TID2013 & 33.2 & 36.8 & 29.2 & 32.2 & 50.4 & 51.1 & 47.0 & 51.1 & \textbf{67.4} & \textbf{67.4} \\
        \bottomrule
    \end{tabular}
    }
\end{table}
\begin{table}
    \centering
    \caption{Mean classification accuracy across 100 train-test random splits on the multiple distortions LIVEMD database for two tasks: distortion type recognition (top); distortion severity and type recognition (bottom).}
    \label{tab:rr-iqa-multidist-reco}
    \resizebox{\columnwidth}{!}{
    \begin{tabular}{l|cccccccccc}
    \toprule
        \multirow{2}{*}{Database} & \multicolumn{10}{c}{Classification accuracy (\%)} \\
         & \multicolumn{2}{c}{AlexNet} & \multicolumn{2}{c}{Inception-v3} & \multicolumn{2}{c}{ResNet-50} & \multicolumn{2}{c}{SqueezeNet-v1.1} & \multicolumn{2}{c}{VGG-16} \\
         & 3-NN & 9-NN & 3-NN & 9-NN & 3-NN & 9-NN & 3-NN & 9-NN & 3-NN & 9-NN \\ \midrule
        LIVEMD & 78.2 & 77.8 & 71.4 & 77.6 & \textbf{90.9} & \textbf{90.9} & 83.7 & 85.1 & 88.6 & 90.2 \\ 
        \midrule
        LIVEMD & 36.2 & 37.8 & 31.9 & 37.5 & 43.7 & \textbf{48.5} & 40.3 & 39.4 & 39.9 & 44.5 \\
        \bottomrule
    \end{tabular}
    }
\end{table}
\section{Conclusions}
In this work, starting from the previous works that have shown how perceptual similarity is an emergent property shared across deep visual representations, we analyze the capability of deep visual representations to intrinsically characterize different types of image distortions. 

In the first experiment we generated a number of synthetically distorted images by applying three mainstream distortion types to the LIVE database and then we analyzed the features extracted from different layers of different deep network architectures. 
From the results of this experiment we observed that a PCA-reduced 2-dimensional representation of the features extracted from a given layer permits to efficiently separate types of distortions in the feature space. 

As a second experiment we evaluated the use of features taken from the layer that better separated image distortions for two different tasks: reduced-reference image quality assessment, and distortion type and severity levels recognition on both single and multiple distortion databases.
The results obtained in this second experiment showed that deep visual representations can be exploited even in an unsupervised way to efficiently recognize various image distortion types and severity levels.
%


\bibliographystyle{model2-names}
\bibliography{refs}

\newpage
\appendix
\section*{Appendix}

\section{Quantitative results}
\label{sec:quant}
\subsection{Analysis}
In this Section we provide more details about the experiments we conducted to verify the effectiveness of deep features for image distortion characterization.
\paragraph{Does the capability of discriminating distortions belong to the network architecture or it depends on the network weights?} In this experiment images are encoded by using deep features extracted from randomly initialized networks. For statistical significance we compute the final distortions separability index for each layer by averaging the scores obtained for 100 network re-initializations. Table \ref{tab:cluster-results-random} depicts the resulting distortions separability indices for each layer of the considered networks. Scores are definitely lower than the ones obtained in the experiment exploiting deep features extracted from the networks pre-trained on ImageNet (see first paragraph of Section 5). This indicates that the degree of separability of images affected by diverse distortions is limited. The ranks of layers per network are completely different from the ones obtained for pre-trained networks, but as it is possible to see the standard deviation of scores among layers is small.
\begin{table*}
    \centering
    \caption{Distortions separability indices for each layer of the considered networks randomly initialized. The indices are calculated by combining three internal cluster validity indices and are included in the interval [0,1], in which the highest scores (in green) indicate a high separability of the types of distortions, while the low scores (in red) represent levels not able to separate the clusters that are formed to discriminate the different distortions.}
    \label{tab:cluster-results-random}
    \resizebox{\textwidth}{!}{
    \begin{tabular}{l|ccccccccccccccc}
        \toprule
        Network & \multicolumn{14}{c}{Layer} \\ \midrule
        \multirow{2}{*}{AlexNet} & \texttt{conv1} & \texttt{conv2} & \texttt{conv3} & \texttt{conv4} & \texttt{conv5} & \multicolumn{9}{c}{} \\ \vspace{.5em}
         & \cellcolor{s5}0.3921 & \cellcolor{s15}0.5111 & \cellcolor{s4}0.5327 & \cellcolor{s3}0.5359 & \cellcolor{s16}0.5351 &  &  &  &  &  &  &  &  & \\
        \multirow{2}{*}{Inception-v3} & \texttt{2a3x3} & \texttt{3b1x1} & \texttt{4a3x3} & \texttt{mixed5b} & \texttt{mixed5c} & \texttt{mixed5d} & \texttt{mixed6a} & \texttt{mixed6b} & \texttt{mixed6c} & \texttt{mixed6d} & \texttt{mixed6e} & \texttt{mixed7a} & \texttt{mixed7b} & \texttt{mixed7c} \\ \vspace{.5em}
         & \cellcolor{s145}0.4075 & \cellcolor{s146}0.4168 & \cellcolor{s60}0.4250 & \cellcolor{s4}0.4285 & \cellcolor{s42}0.4353 & \cellcolor{s3}0.4415 & \cellcolor{s147}0.4379 & \cellcolor{s148}0.4333 & \cellcolor{s149}0.4372 & \cellcolor{s86}0.4272 & \cellcolor{s71}0.4284 & \cellcolor{s150}0.4223 & \cellcolor{s123}0.4292 & \cellcolor{s5}0.3708 \\
        \multirow{2}{*}{ResNet-50} & \texttt{conv1} & \texttt{layer1} & \texttt{layer2} & \texttt{layer3} & \texttt{layer4} &  &  &  &  &  &  &  &  & \\ \vspace{.5em}
         & \cellcolor{s5}0.3943 & \cellcolor{s98}0.5418 & \cellcolor{s14}0.5530 & \cellcolor{s4}0.5489 & \cellcolor{s3}0.5565 &  &  &  &  &  &  &  &  & \\
        \multirow{2}{*}{SqueezeNet-v1.1} & \texttt{conv1} & \texttt{fire1} & \texttt{fire2} & \texttt{fire3} & \texttt{fire4} & \texttt{fire5} & \texttt{fire6} & \texttt{fire7} & \texttt{fire8} &  &  &  &  & \\ \vspace{.5em}
        & \cellcolor{s5}0.3928 & \cellcolor{s185}0.5159 & \cellcolor{s122}0.5238 & \cellcolor{s71}0.5703 & \cellcolor{s4}0.5712 & \cellcolor{s186}0.5809 & \cellcolor{s187}0.5796 & \cellcolor{s9}0.5928 & \cellcolor{s3}0.6621 &  &  &  &  & \\
        \multirow{2}{*}{VGG-16} & \texttt{conv11} & \texttt{conv12} & \texttt{conv21} & \texttt{conv22} & \texttt{conv31} & \texttt{conv32} & \texttt{conv33} & \texttt{conv41} & \texttt{conv42} & \texttt{conv43} & \texttt{conv51} & \texttt{conv52} & \texttt{conv53} & \\
         & \cellcolor{s5}0.3944 & \cellcolor{s59}0.4069 & \cellcolor{s68}0.5058 & \cellcolor{s69}0.5088 & \cellcolor{s70}0.5577 & \cellcolor{s4}0.5621 & \cellcolor{s71}0.5618 & \cellcolor{s72}0.5745 & \cellcolor{s73}0.5780 & \cellcolor{s74}0.5765 & \cellcolor{s75}0.5823 & \cellcolor{s76}0.5850 & \cellcolor{s3}0.5853 & \\ \bottomrule
    \end{tabular}}
\end{table*}
\paragraph{Can the removal of semantics from the visual representation of the image emphasize the distortion encoding?} In this experiment we want to understand if the separability property of clusters in the deep feature space is influenced by the content of the images. For doing that, we subtract in channel dimension the vector of features, of size $1 \times c_l$, of the reference image to the feature vector of the distorted image. In Table \ref{tab:cluster-results-nosematics} we report the results of this experiment, which indicate that the semantics does not affect the rank of the layers for AlexNet, VGG-16, as well as for ResNet-50 which, however, presents an increase in the distortions separability index for the best layer (i.e. \texttt{layer1}). This may be due to the presence of skip connections that carry low-level information from previous layers. Both Inception-v3 and SqueezeNet-v1.1 obtain a different rank of layers with also lower scores for the distortions separability index.
\begin{table*}
    \centering
    \caption{\textbf{Distortions separability indices for each layer of the considered networks pre-trained on ImageNet without semantics.} To isolate the signal encoding the semantics we subtract the deep features encoding the reference image from the deep features for the distorted one. The indices are calculated by combining three internal cluster validity indices and are included in the interval [0,1], in which the highest scores (in green) indicate a high separability of the types of distortions, while the low scores (in red) represent levels not able to separate the clusters that are formed to discriminate the different distortions.}
    \label{tab:cluster-results-nosematics}
    \resizebox{\textwidth}{!}{
    \begin{tabular}{l|ccccccccccccccc}
        \toprule
        Network & \multicolumn{14}{c}{Layer} \\ \midrule
        \multirow{2}{*}{AlexNet} & \texttt{conv1} & \texttt{conv2} & \texttt{conv3} & \texttt{conv4} & \texttt{conv5} & \multicolumn{9}{c}{} \\ \vspace{.5em}
         & \cellcolor{s11}0.5360 & \cellcolor{s3}0.5810 & \cellcolor{s4}0.5183 & \cellcolor{s5}0.4544 & \cellcolor{s12}0.4622 &  &  &  &  &  &  &  &  & \\
        \multirow{2}{*}{Inception-v3} & \texttt{2a3x3} & \texttt{3b1x1} & \texttt{4a3x3} & \texttt{mixed5b} & \texttt{mixed5c} & \texttt{mixed5d} & \texttt{mixed6a} & \texttt{mixed6b} & \texttt{mixed6c} & \texttt{mixed6d} & \texttt{mixed6e} & \texttt{mixed7a} & \texttt{mixed7b} & \texttt{mixed7c} \\ \vspace{.5em}
         & \cellcolor{s5}0.3428 & \cellcolor{s126}0.4052 & \cellcolor{s127}0.4161 & \cellcolor{s128}0.3938 & \cellcolor{s29}0.3983 & \cellcolor{s129}0.4009 & \cellcolor{s130}0.4375 & \cellcolor{s131}0.4675 & \cellcolor{s132}0.4608 & \cellcolor{s66}0.4713 & \cellcolor{s133}0.4550 & \cellcolor{s134}0.4540 & \cellcolor{s135}0.5579 & \cellcolor{s3}0.5619 \\
        \multirow{2}{*}{ResNet-50} & \texttt{conv1} & \texttt{layer1} & \texttt{layer2} & \texttt{layer3} & \texttt{layer4} &  &  &  &  &  &  &  &  & \\ \vspace{.5em}
         & \cellcolor{s5}0.3935 & \cellcolor{s3}0.7791 & \cellcolor{s95}0.6216 & \cellcolor{s4}0.4686 & \cellcolor{s4}0.4686 &  &  &  &  &  &  &  &  & \\
        \multirow{2}{*}{SqueezeNet-v1.1} & \texttt{conv1} & \texttt{fire1} & \texttt{fire2} & \texttt{fire3} & \texttt{fire4} & \texttt{fire5} & \texttt{fire6} & \texttt{fire7} & \texttt{fire8} &  &  &  &  & \\ \vspace{.5em}
        & \cellcolor{s5}0.4904 & \cellcolor{s97}0.6268 & \cellcolor{s179}0.6547 & \cellcolor{s4}0.6442 & \cellcolor{s180}0.6655 & \cellcolor{s7}0.6596 & \cellcolor{s160}0.5942 & \cellcolor{s10}0.6121 & \cellcolor{s3}0.7132 &  &  &  &  & \\
        \multirow{2}{*}{VGG-16} & \texttt{conv11} & \texttt{conv12} & \texttt{conv21} & \texttt{conv22} & \texttt{conv31} & \texttt{conv32} & \texttt{conv33} & \texttt{conv41} & \texttt{conv42} & \texttt{conv43} & \texttt{conv51} & \texttt{conv52} & \texttt{conv53} & \\
         & \cellcolor{s49}0.5040 & \cellcolor{s50}0.6258 & \cellcolor{s51}0.6459 & \cellcolor{s3}0.6905 & \cellcolor{s52}0.6480 & \cellcolor{s53}0.6200 & \cellcolor{s54}0.5331 & \cellcolor{s55}0.5735 & \cellcolor{s4}0.5452 & \cellcolor{s56}0.4962 & \cellcolor{s57}0.4494 & \cellcolor{s58}0.4133 & \cellcolor{s5}0.3889 & \\ \bottomrule
    \end{tabular}}
\end{table*}
\subsection{Applications}
In this Section we provide a more detailed analysis of the correlation between the estimated quality scores and the ground-truth ones.
\paragraph{Reduced-Reference Image Quality Assessment} In Table \ref{tab:iqa-results-one-distortion-at-time} we report results for the reduced-reference image quality assessment task. We show  correlations estimated for each image distortion separately, i.e. additive white Gaussian noise (AWGN), Gaussian blur and JPEG compression, on each of the four image quality databases considered (CSIQ, LIVE, TID2008, and TID2013). 

\begin{table}
\centering
\caption{Median and mean SROCC and PLCC values across 100 train-test random splits on different types of distortions on CSIQ, LIVE, TID2008, and TID2013 databases for the considered networks.}
\label{tab:iqa-results-one-distortion-at-time}
\resizebox{\columnwidth}{!}{
\begin{tabular}{lll|cccccc}
\toprule
\multirow{2}{*}{Database} & \multirow{2}{*}{Network} & \multirow{2}{*}{Dist. type} & \multicolumn{2}{c}{Median} & \multicolumn{2}{c}{Mean}\\
& & & SROCC & LCC &  SROCC & LCC \\ \midrule
\multirow{17}{*}{CSIQ} & AlexNet & \multirow{5}{*}{AWGN} & \textbf{0.9533} & \textbf{0.9527} & \textbf{0.9532} & \textbf{0.9523} \\
 & Inception-v3 & & 0.9168 & 0.9174 & 0.9110 & 0.9103 \\
 & ResNet-50 & & 0.9359 & 0.9374 & 0.9347 & 0.9354 \\
 & SqueezeNet-v1.1 & & 0.9040 & 0.8772 & 0.8976 & 0.8733 \\
 & VGG-16 & & 0.9416 & 0.9432 & 0.9394 & 0.9410 \\
& & & & \\
 & AlexNet & \multirow{5}{*}{GBlur} & 0.8982 & 0.8540 & 0.8968 & 0.8568 \\
 & Inception-v3 & & \textbf{0.9702} & \textbf{0.9679} & \textbf{0.9654} & \textbf{0.9630} \\
 & ResNet-50 & & 0.9617 &  0.9614 & 0.9572 & 0.9555 \\
 & SqueezeNet-v1.1 & & 0.9519 & 0.9366 & 0.9501 & 0.9353 \\
 & VGG-16 & & 0.8977 & 0.8625 & 0.8837 & 0.8444 \\
& & & & \\
 & AlexNet & \multirow{5}{*}{JPEG} & 0.9306 & 0.9462 & 0.9235 & 0.9324 \\
 & Inception-v3 & & \textbf{0.9563} & 0.9470 & \textbf{0.9493} & 0.9413 \\
 & ResNet-50 & & 0.9497 & \textbf{0.9518} & 0.9434 & \textbf{0.9461} \\
 & SqueezeNet-v1.1 & & 0.9153 & 0.9373 & 0.9116 & 0.9324 \\
 & VGG-16 & & 0.9221 & 0.9245 & 0.9136 & 0.9101 \\ \midrule
\multirow{17}{*}{LIVE} & AlexNet & \multirow{5}{*}{AWGN} & \textbf{0.9844} & 0.9566 & \textbf{0.9838} & 0.9561 \\
 & Inception-v3 & & 0.9266 & 0.9262 & 0.9206 & 0.9210 \\
 & ResNet-50 & & 0.9742 & \textbf{0.9822} & 0.9727 & \textbf{0.9805} \\
 & SqueezeNet-v1.1 & & 0.9506 & 0.9531 & 0.9427 & 0.9331 \\
 & VGG-16 & & 0.9782 & 0.9641 & 0.9777 & 0.9637 \\
& & & & \\
 & AlexNet & \multirow{5}{*}{GBlur} & 0.8598 & 0.8501 & 0.8325 & 0.8053 \\
 & Inception-v3 & & 0.9377 & \textbf{0.9433} & 0.9368 & \textbf{0.9400} \\ 
 & ResNet-50 & & \textbf{0.9399} & 0.9384 & \textbf{0.9377} & 0.9323 \\
 & SqueezeNet-v1.1 & & 0.9190 & 0.8940 & 0.9123 & 0.8929 \\
 & VGG-16 & & 0.8407 & 0.8085 & 0.8355 & 0.8000 \\
& & & & \\
& AlexNet & \multirow{5}{*}{JPEG} & 0.9518 & \textbf{0.9623} & 0.9455 & 0.9462 \\
 & Inception-v3 & & 0.9325 & 0.9308 & 0.9301 & 0.9289 \\
 & ResNet-50 & & \textbf{0.9546} & 0.9563 & \textbf{0.9458} & \textbf{0.9473} \\
 & SqueezeNet-v1.1 & & 0.9254 & 0.9136 & 0.9207 & 0.9130 \\
 & VGG-16 & & 0.9149 & 0.9163 & 0.9079 & 0.9091 \\ \midrule
\multirow{17}{*}{TID2008} & AlexNet & \multirow{5}{*}{AWGN} & 0.9395 & 0.9387 & 0.9358 & 0.9356 \\
 & Inception-v3 & & 0.7549 & 0.7511 & 0.7500 & 0.7495 \\
 & ResNet-50 & & 0.9278 & 0.9237 & 0.9263 & 0.9206 \\
 & SqueezeNet-v1.1 & & 0.7714 & 0.7379 & 0.7736 & 0.7456 \\
 & VGG-16 & & \textbf{0.9485} & \textbf{0.9470} & \textbf{0.9407} & \textbf{0.9375} \\
& & & & \\
 & AlexNet & \multirow{5}{*}{GBlur} & 0.8977 & 0.8705 & 0.8463 & 0.8058 \\
 & Inception-v3 & & \textbf{0.9639} & \textbf{0.9633} & \textbf{0.9520} & \textbf{0.9502} \\
 & ResNet-50 & & 0.9364 & 0.9372 & 0.9306 & 0.9339 \\
 & SqueezeNet-v1.1 & & 0.8722 & 0.8666 & 0.8637 & 0.8533 \\
 & VGG-16 & & 0.9263 & 0.9152 & 0.8692 & 0.8292 \\
& & & & \\
 & AlexNet & \multirow{5}{*}{JPEG} & \textbf{0.9459} & 0.9576 & \textbf{0.9449} & \textbf{0.9496} \\
 & Inception-v3 & & 0.8845 & 0.8998 & 0.8701 & 0.8833 \\
 & ResNet-50 & & 0.9308 & \textbf{0.9592} & 0.9285 & 0.9485 \\
 & SqueezeNet-v1.1 & & 0.8647 & 0.8742 & 0.8352 & 0.8367 \\
 & VGG-16 & & 0.9289 & 0.9452 & 0.9217 & 0.9187 \\ \midrule
\multirow{17}{*}{TID2013} & AlexNet & \multirow{5}{*}{AWGN} & \textbf{0.9521} & 0.9489 & \textbf{0.9470} & \textbf{0.9469} \\
 & Inception-v3 & & 0.8332 & 0.8291 & 0.8194 & 0.8147 \\
 & ResNet-50 & & 0.9300 & 0.9317 & 0.9257 & 0.9268 \\
 & SqueezeNet-v1.1 & & 0.8246 & 0.8111 & 0.8269 & 0.8109 \\
 & VGG-16 & & 0.9434 & \textbf{0.9495} & 0.9380 & 0.9438 \\
& & & & \\
 & AlexNet & \multirow{5}{*}{GBlur} & 0.9446 & 0.9070 & 0.9163 & 0.8779 \\
 & Inception-v3 & & \textbf{0.9646} & \textbf{0.9629} & \textbf{0.9584} & \textbf{0.9565} \\
 & ResNet-50 & & 0.9533 & 0.9593 & 0.9518 & 0.9563 \\
 & SqueezeNet-v1.1 & & 0.9069 & 0.8992 & 0.8966 & 0.8872 \\
 & VGG-16 & & 0.9500 & 0.9555 & 0.9178 & 0.8893 \\
& & & & \\
 & AlexNet & \multirow{5}{*}{JPEG} & \textbf{0.9615} & \textbf{0.9752} & \textbf{0.9615} & \textbf{0.9688} \\
 & Inception-v3 & & 0.9136 & 0.9249 & 0.8949 & 0.9074 \\
 & ResNet-50 & & 0.9377 & 0.9592 & 0.9327 & 0.9508 \\
 & SqueezeNet-v1.1 & & 0.8932 & 0.9075 & 0.8667 & 0.8623 \\
 & VGG-16 & & 0.9500 & 0.9557 & 0.9460 & 0.9430 \\ \bottomrule
\end{tabular}}
\end{table}

\end{document}